\documentclass[letterpaper, 10 pt, conference]{ieeeconf}  % Comment this line out if you need a4paper

\IEEEoverridecommandlockouts                              % This command is only needed if 
\usepackage{amsmath}
\usepackage{amssymb}    
\usepackage{amsfonts}  
\usepackage{graphicx}
\usepackage{booktabs}
\usepackage{multirow}
\usepackage{colortbl}
\usepackage{xcolor}
\usepackage{array}
\usepackage{multirow}
\usepackage{caption}
\usepackage[table]{xcolor}
\usepackage{caption}
\usepackage[table]{xcolor}

\title{\LARGE \bf
Agile-VLA: Few-Shot Industrial Pose Rectification via Implicit Affordance Anchoring
}

\author{
Teng Yan$^{1}$,
Zhengyang Pei$^{1}$,
Chengyu Shi$^{1}$,
Yue Yu$^{1}$,
Yikun Chen$^{1}$,
Zilong Zhu$^{1}$, \\
Zelin Fang$^{1}$,
Kaile Guo$^{1}$,
Zihang Wang$^{1}$,
Peigen Tian$^{1}$,
and Bingzhuo Zhong$^{1}$\thanks{Corresponding author.}%
\\
$^{1}$The Hong Kong University of Science and Technology (Guangzhou)
\\
{\tt\small tyan497@connect.hkust-gz.edu.cn, bingzhuoz@hkust-gz.edu.cn
}
}

\begin{document}

\maketitle
\thispagestyle{empty}
\pagestyle{empty}

%%%%%%%%%%%%%%%%%%%%%%%%%%%%%%%%%%%%%%%%%%%%%%%%%%%%%%%%%%%%%%%%%%%%%%%%%%%%%%%%
\begin{abstract}

Deploying Vision–Language–Action (VLA) models on resource-constrained edge platforms encounters a fundamental conflict between high-latency semantic inference and the high-frequency control required for dynamic manipulation. To address the challenge, this paper presents \textbf{Agile-VLA}, a hierarchical framework designed for industrial pose reorientation tasks on edge devices such as the NVIDIA Jetson Orin Nano. The core innovation is an Implicit Affordance Anchoring mechanism that directly maps geometric visual cues, specifically centroid and rim keypoint anchors, into structured parametric action primitives, thereby substantially reducing reliance on high-latency semantic inference during closed-loop control. By decoupling perception (10 Hz) from control (50 Hz) via an asynchronous dual-stream architecture, the system effectively mitigates the frequency mismatch inherent in edge-based robot learning. Experimental results on a standard 6-DoF manipulator demonstrate that Agile-VLA achieves robust rectification of complex, irregular workpieces using only 5-shot demonstrations through extrinsic dexterity.

\end{abstract}

%%%%%%%%%%%%%%%%%%%%%%%%%%%%%%%%%%%%%%%%%%%%%%%%%%%%%%%%%%%%%%%%%%%%%%%%%%%%%%%%
\section{INTRODUCTION}
The transition toward flexible manufacturing requires robotic systems to operate stably in high-mix, low-volume production environments involving structurally complex components, without tedious manual reprogramming. In industrial assembly lines, many thin-shell or irregular workpieces, such as stamped parts, castings, or composite structural components, exhibit distinct functional surface asymmetries. Their spatial orientation directly affects downstream operations, including assembly, quality inspection, and subsequent handling. Consequently, robots must move beyond mere shape recognition and instead perform function-oriented pose adjustment grounded in structural semantics.

\begin{figure}
    \centering
    \includegraphics[width=1.0\linewidth]{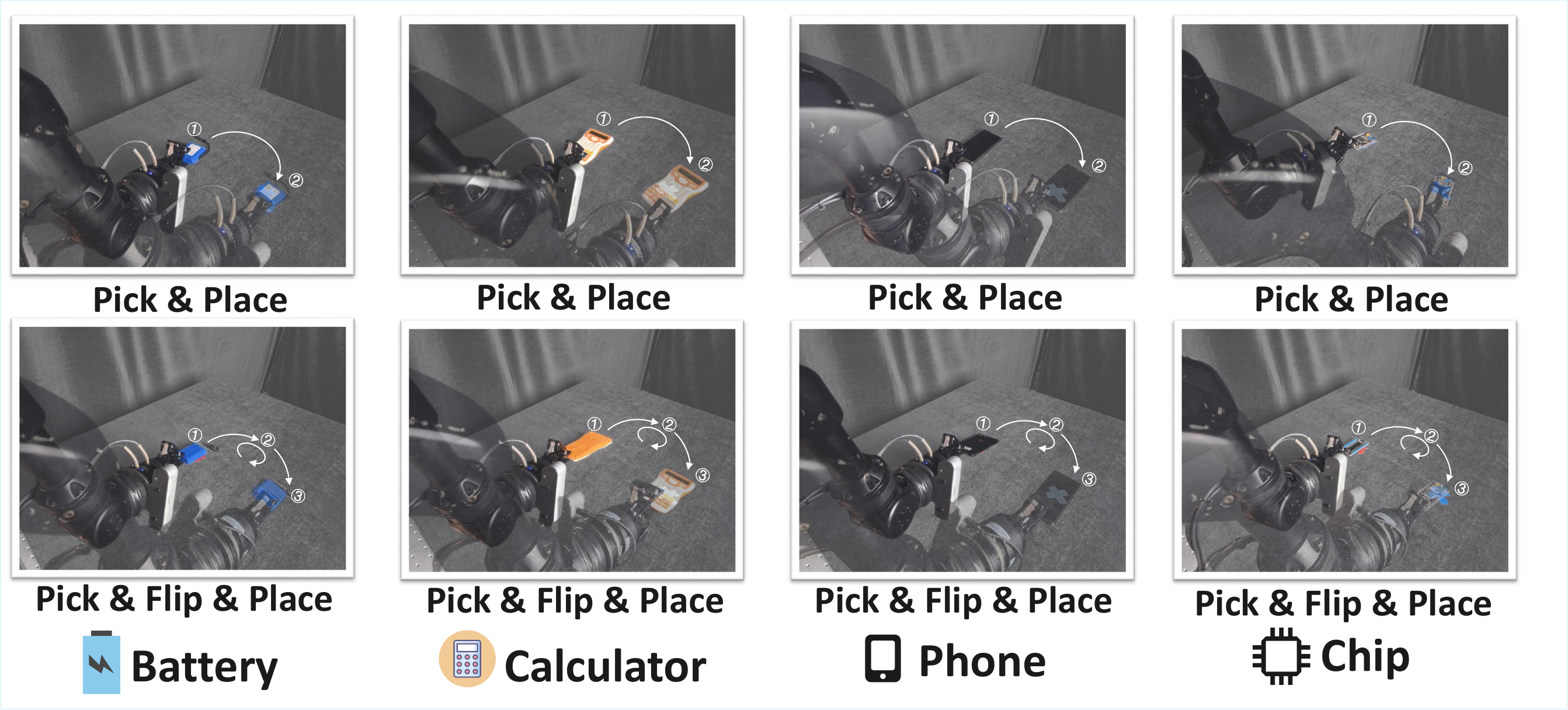}
    \caption{\textbf{Visualization of Agile-VLA task primitives across diverse industrial workpieces.} The top row illustrates Stable Pick and Place operations guided by Stability Anchoring, where the grasp point is optimized toward the geometric centroid to minimize gravitational torque. The bottom row demonstrates Pick, Flip and Place sequences leveraging Pivot Anchoring; here, the anchor point is constrained to the object boundary to maximize the lever arm for extrinsic dexterity. The framework is validated on four irregular objects: (a) Battery, (b) Calculator, (c) Phone, and (d) Chip.}
    \label{fig:placeholder}
\end{figure}

Traditional automation typically depends on rigid, customized fixtures to manipulate irregular parts. Although upgrading to high-DoF manipulators or expensive dexterous hands offers greater flexibility, such hardware modifications prohibitively increase both deployment costs and maintenance complexity. In practice, the majority of industrial assembly lines remain constrained to standard 6-DoF arms with simple parallel grippers. Under these hardware limitations, leveraging extrinsic dexterity provides a compelling solution. It empowers standard 6-DoF manipulators to exploit environmental contacts—such as tabletop edges—as external pivots. This mechanism facilitates complex flipping and reorientation maneuvers that would otherwise necessitate fully actuated dexterous hands. And the core task primitives enabled by this approach are illustrated in Fig.~\ref{fig:placeholder}.

Meanwhile, Vision-Language-Action (VLA) models offer new possibilities for cross-task generalization. However, implementing fine-grained manipulation on edge devices, such as the NVIDIA Jetson Orin Nano, faces dual challenges. The primary challenge lies in the computation-control frequency gap. On edge hardware, large VLA models typically exhibit inference latencies $T_{inf}$ between $100\,\text{ms}$ and $500\,\text{ms}$, while flipping operations involving physical contact require control frequencies $f_{ctrl} > 30\,\text{Hz}$. Conventional synchronous end-to-end policies $\pi(a_t|I_t)$ can lead to severe system instability under such latency constraints. Furthermore, purely semantic descriptions often fail to capture the fine-grained geometric features required for contact-intensive manipulation.

\begin{figure*}[t]
    \centering
    \includegraphics[width=1.0\linewidth]{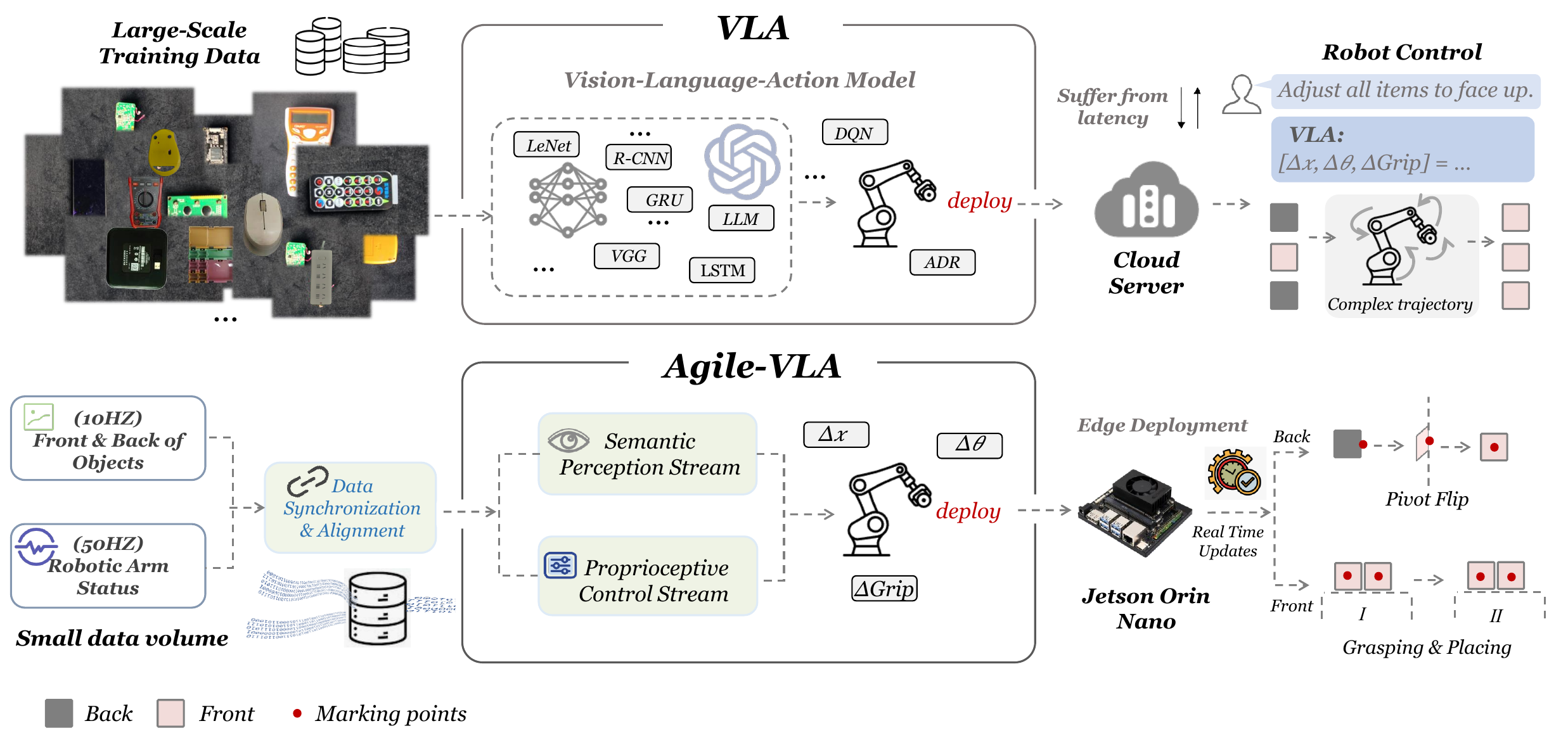}
    \caption{\textbf{System Overview.} Compared to conventional large-scale VLA models that suffer from high cloud latency, Agile-VLA achieves real-time, few-shot pose reorientation directly on an edge device (NVIDIA Jetson Orin Nano). By decoupling high-level semantic reasoning from low-level control, the framework enables agile manipulation of complex industrial components using only a minimal number of demonstration samples.}
    \label{fig:system}
\end{figure*}

To bridge these gaps, this paper presents \textbf{Agile-VLA}, an industrial-grade framework for agile perception and manipulation. And the system overview is showed in Fig \ref{fig:system}. The key contributions include:

\begin{itemize}
    \item \textbf{Implicit Affordance Anchoring.} Manipulation logic is directly encoded into geometric keypoint locations. By identifying asymmetric functional features, the model dynamically generates action anchors, specifically, targeting the centroid of functional surfaces for secure grasping and the rim of non-functional surfaces for execution of flipping primitives.
    \item \textbf{Asynchronous Dual-Stream Architecture.} Perception (10 Hz) and control (50 Hz) are temporally decoupled, maintaining real-time stability on edge hardware despite inference latency.
    \item \textbf{Extrinsic Dexterity Primitives.} The workbench surface is utilized as an external pivot to compensate for the kinematic constraints of standard 6-DoF manipulators. This approach enables the robust rectification of complex, non-regular workpieces without the need for expensive multi-fingered dexterous hands.
\end{itemize}

\section{Related Work}

\subsection{Vision-Language-Action (VLA) Models and Edge Efficiency}
In recent years, the conversion of large-scale Vision-Language Models (VLMs) into embodied action policies has become a major research focus. RT-1 \cite{c1} and RT-2 \cite{c2} established the preliminary framework for semantic generalization through internet-scale knowledge. And the Open X-Embodiment project, through RT-X \cite{c3}, demonstrated the potential of massive cross platform datasets to enhance robotic universality. Additionally, multi-modality approaches such as TextGCD \cite{c21} have demonstrated that leveraging textual knowledge alongside visual cues can significantly improve the discovery and clustering of novel categories, highlighting the potential of cross-modality co-teaching in embodied perception tasks. According to the latest surveys on VLA systems \cite{c4}, the release of models such as Octo \cite{c5}, pi0 \cite{c8} and Dita \cite{c19} indicates that robotic foundation models are evolving toward high-frequency and continuous action generation.

Despite these advances, high performance in such models is typically supported by large parameter counts and complex autoregressive generation mechanisms. Recent empirical evidence suggests that even quantized versions of OpenVLA struggle to exceed an inference frequency of 5\,Hz on edge hardware \cite{c4}. To address this bottleneck, EdgeVLA (EVLA, 2025) \cite{c6} removes the autoregressive requirement in action prediction, achieving a sevenfold acceleration in inference. Further progress was marked by NanoVLA (2025) \cite{c7}, which utilized delayed fusion for vision-language reasoning and a decoupled architecture to realize a 52-fold speedup on the Jetson Orin Nano platform. In parallel, diffusion-based formulations have been explored to enhance robustness and efficiency in perception modules. For instance, RSDNet \cite{c20} leverages a detachable latent diffusion framework to learn denoising in feature space and enables single-step inference, reducing the computational overhead typically associated with iterative diffusion processes. Agile-VLA extends this line of development by adopting an asynchronous hierarchical design that addresses the frequency mismatch between semantic perception and high-frequency control in edge-deployed robotic systems.

\subsection{Perception via Keypoint and Implicit Affordance}
Representation methods based on geometric keypoints have demonstrated superior performance in enhancing data efficiency and generalization. While early deep learning frameworks like AffordanceNet \cite{c14} focused on explicit semantic segmentation of functional regions, kPAM \cite{c11} demonstrates that sparse semantic keypoints can effectively capture object affordances closely associated with geometric structures, while Transporter Networks \cite{c17} leverage spatial equivariance properties to significantly improve the data efficiency and generalization capability of imitation learning. Recent advances such as Any-point \cite{c12} demonstrate the feasibility of general-purpose manipulation through the tracking of arbitrary keypoint trajectories within image space. Furthermore, ManiGaussian \cite{c13} introduced 3D Gaussian Splatting techniques to enhance grasp robustness through precise dynamic scene representation.

The key contribution of Agile-VLA lies in the formulation of Implicit Affordance Anchoring. Unlike conventional keypoint methods that serve primarily as spatial references, this approach implicitly encodes operational logic within coordinate locations through a specific labeling protocol, such as centroid versus rim labeling. This reduced representation lowers computational demands on edge hardware while enabling action switching for novel workpieces with as few as five demonstrations. To further address the challenges of identifying and categorizing such novel instances in unconstrained environments, recent advancements in open-world learning, such as the CAL framework \cite{c22}, provide a robust reference. By balancing model confidence and automatically estimating the number of unknown categories through Dynamic Prototype Pruning (DPP), these methods mitigate the bias toward known types, offering potential for VLA systems to autonomously perceive the boundaries of their task-specific knowledge.

\subsection{Asynchronous Architectures and Extrinsic Dexterity}
An alternative trajectory to mitigate the discrepancy between large-model latency and real-time control adopts asynchronous control architectures. VLA-Cache \cite{c9} reduces redundant computation through adaptive token caching, and the acceleration of VLA execution through parallel decoding and action chunking was explored by Song et al. \cite{c10}. More recently, Real-Time Chunking \cite{c18} demonstrated the advantages of asynchronous flow matching in generating smooth action trajectories at 50\,Hz. The coherence of manipulation sequences was further enhanced by Chain-of-Action (2026) \cite{c15} through the integration of autoregressive trajectory modeling.

At the physical execution level, extrinsic dexterity plays an important role in compensating for the limited degrees of freedom of simple parallel grippers. Chavan-Dafle et al. provide a systematic analysis of quasi-static reorientation using tabletop contact as a pivot constraint. \cite{c16}. 
Given that 6-DoF manipulators are prone to kinematic singularities, Agile-VLA constructs an action primitive library grounded in environmental constraints. Through vision-based rim anchoring derived from optical flow prediction, the system leverages the tabletop as an external pivot to accomplish flipping motions. This strategy preserves deterministic pose reorientation under computational constraints.

\section{Methodology}

\subsection{Problem Formulation: Edge-Constrained Hybrid Control}
The pose reorientation task on an industrial assembly line is formulated as an edge-constrained hybrid dynamical system. In this setting, $\mathcal{C}_{edge}$ denotes dual physical constraints imposed by the deployment on embedded hardware (e.g., Jetson Orin Nano), namely computational and memory restrictions that prohibit full fine-tuning of models with more than 7B parameters, together with a temporal mismatch characterized by visual inference latency $T_{inf}$ that substantially exceeds the control period $\Delta t_{ctrl}$ i.e., $T_{inf} \gg 1/f_{ctrl}$.

The system state is defined as:
\begin{equation}
    s_t = \{I_t, q_t, \dot{q}_t\}
\end{equation}
where $I_t \in \mathbb{R}^{H \times W \times 3}$ represents high-dimensional visual observations, and $q_t, \dot{q}_t \in \mathbb{R}^6$ denote the joint angles and velocities of the robotic manipulator, respectively. 
The control objective is to learn an optimal policy $\pi^*$ that that achieves pose reorientation from a random initial pose $SO(3)_{init}$ to the target pose $SO(3)_{goal}$ under the constraint $\mathcal{C}_{edge}$:

\begin{equation}
\pi^* = \arg\min_{\pi} \mathbb{E} \left[ \sum_{t=0}^{T} \| \mathcal{T}(s_t) \ominus SO(3)_{goal} \|^2 \right]
\end{equation}
\begin{equation}
\text{s.t.} \quad T_{inf}(\pi) \approx \Delta t_{ctrl} \quad (\text{Agility Constraint})
\end{equation}

This constraint formally characterizes the requirement that architectural design must mitigate the closed-loop instability caused by the mismatch $T_{inf} \gg 1/f_{ctrl}$. Such instability constitutes the fundamental systems-level challenge in edge-deployed pose reorientation (also referred to as pose rectification), thereby motivating the introduction of an asynchronous architecture.

\begin{figure}
    \centering
    \includegraphics[width=1.0\linewidth]{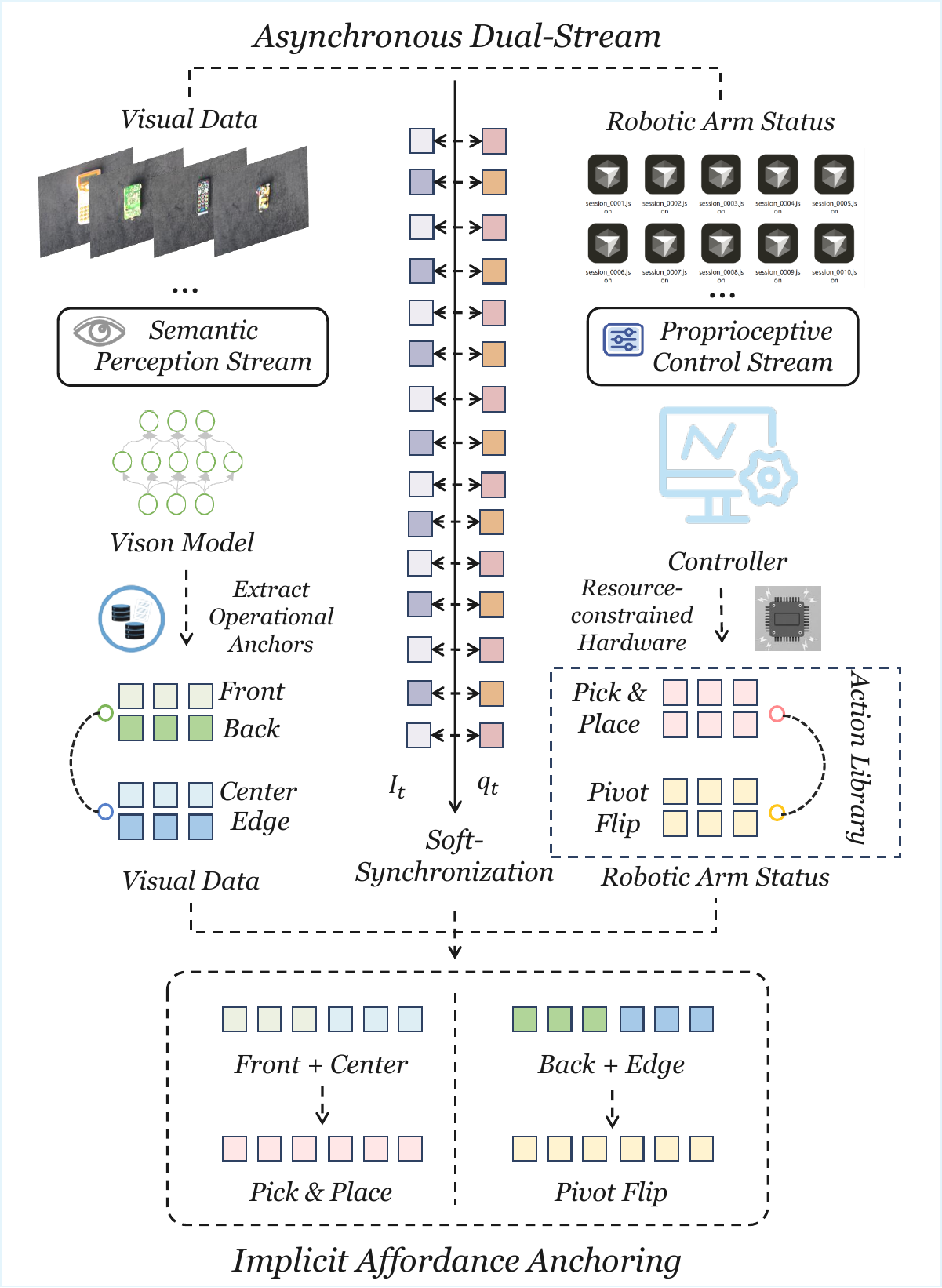}
    \caption{\textbf{Asynchronous Dual-Stream Framework.} The semantic perception stream (10 Hz) is responsible for extracting manipulation anchors, while the proprioceptive control stream (50 Hz) executes parameterized action primitives. A timestamp-based soft synchronization protocol ensures strict alignment between the visual and kinesthetic streams, thereby enabling high-fidelity few-shot adaptation on resource-constrained hardware.}
    \label{fig:framework}
\end{figure}

\subsection{Architecture: Asynchronous Dual-Stream Decoupling}
\begin{figure*}
    \centering
    \includegraphics[width=1.0\linewidth]{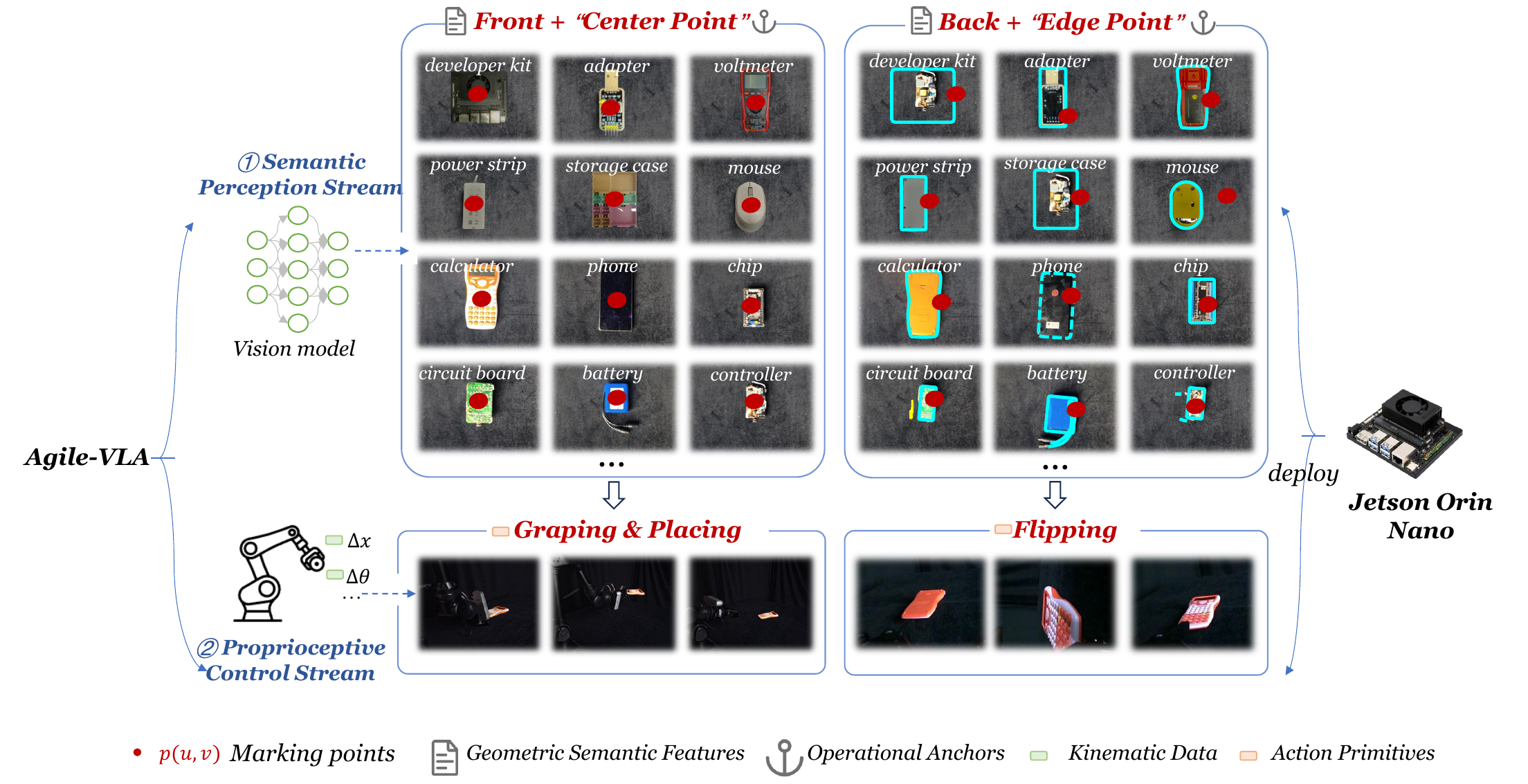}
    \caption{\textbf{Implicit Affordance Anchoring Protocol.} The system maps visual cues to specific action primitives according to the object’s topological state. Front-facing objects are anchored at their geometric centroid to trigger stable, guidance-oriented grasping, whereas back-facing objects are anchored at the rim to initiate extrinsic-dexterity-based flipping, thereby effectively leveraging environmental constraints to overcome kinematic limitations.}
    \label{fig:Mechanism}
\end{figure*}

As shown in Fig. \ref{fig:Mechanism} Agile-VLA achieves computational agility through an asynchronous dual-stream architecture, which fundamentally decouples high-level semantic perception from real-time control, thereby enabling efficient semantic reasoning without compromising low-latency control execution.:

\begin{itemize}
    \item \textbf{Semantic Perception Stream (10Hz).} A TensorRT-accelerated and quantized vision model extracts geometric-semantic features and operational anchors, denoted as $\mathcal{O}_k = \mathcal{F}_{vis}(I_{t_k})$.
    \item \textbf{Proprioceptive Control Stream (50Hz).} Real-time motion primitives are executed via the ServoJ interface, while synchronized proprioceptive data are streamed in JSON format.
\end{itemize}

To bridge the frequency gap between perception (10Hz) and control (50Hz), a cubic spline interpolation kernel $\mathcal{S}$ is introduced within the controller. For any control instant $t \in [t_k, t_{k+1}]$, the continuous control trajectory $q_{cmd}(t)$ is reconstructed from a sequence of discrete anchors $\{ \mathbf{p}_k \}$:
\begin{equation}
q_{cmd}(t) = \mathcal{S}(\{ \mathbf{p}_k, \mathbf{p}_{k-1}, \mathbf{p}_{k-2}, \mathbf{p}_{k-3} \}; t)
\end{equation}
This interpolation operator ensures $C^2$ continuity in joint space (continuity in position, velocity, and acceleration), thereby generating smooth high-frequency control commands guided by sparse visual inputs.

Furthermore, a timestamp-based soft-synchronization protocol is employed to maintain multimodal data consistency. Defining the visual frame timestamp as $T(I_{t_v})$ and the robot state timestamp as $T(q_{t_c})$, the alignment must satisfy the following temporal constraint:
\begin{equation}
| T(I_{t_v}) - T(q_{t_c}) | < \epsilon_{sync}, \quad \epsilon_{sync} = 10\,\text{ms}
\end{equation}

\subsection{Implicit Affordance Anchoring}
As illustrated in Fig. \ref{fig:framework} the central theoretical contribution lies in Implicit Affordance Anchoring, a mechanism that encodes complex manipulation logic into the spatial representations of visual keypoints through geometric semantic prompts. The VLA output is defined as a tuple 
\begin{equation}
    \mathcal{O} = \{c, \mathbf{p}\}
\end{equation}

where $c \in \{ \text{Front}, \text{Back} \}$ denotes the affordance category and $\mathbf{p} \in \mathbb{R}^2$ represents the corresponding geometric anchor. 

Through a mapping function $\Phi: \mathcal{O} \rightarrow \mathcal{A}$, the system selects action primitives based on the principle of geometric energy minimization:

\textbf{Stability Anchoring (Front + “Center Point”).} When the workpiece is in the front-facing state ($c = \text{Front}$), the anchor $\mathbf{p}$ converges to the geometric centroid $\mathbf{c}_{geo}$. This implicitly encodes a stability prior aimed at minimizing gravitational torque deviation during grasping:

\begin{equation}
\begin{aligned}
\mathbf{p}^* &= \arg\min_{\mathbf{p}} \| \mathbf{p} -\mathbf{c}_{geo} \|_2 \\
&\implies \text{Trigger: Stable Pick-and-Place}
\end{aligned}
\end{equation}
This optimization objective ensures the grasp point aligns with the principal axes of inertia, thereby improving force distribution and maximizing grasp stability.

\textbf{Pivot Anchoring (Back + “Edge Point”).} When the workpiece is back-facing ($c = \text{Back}$), the anchor $\mathbf{p}$ is constrained to the object boundary $\partial \Omega$. This design is intended to maximize the lever arm $L(\mathbf{p})$ relative to the centroid:

\begin{equation}
\begin{aligned}
    \mathbf{p}^* &= \arg\max_{\mathbf{p} \in \partial \Omega} \| \mathbf{p} - \mathbf{c}_{geo} \|_2 \\
    &\implies \text{Trigger: Pivot-Flip}
\end{aligned}
\end{equation}
The selection of the edge anchor is not stochastic but is strategically chosen to gain optimal mechanical advantage for the subsequent flipping operation.

\subsection{Parameterized Action Primitives via Extrinsic Dexterity}
To address the kinematic limitations of a 6-DoF manipulator in pose reorientation, a library of atomic action primitives is constructed based on extrinsic dexterity. By explicitly leveraging environmental constraints, the design establishes a constrained contact dynamics model that compensates for the limited intrinsic degrees of freedom of a parallel gripper.

\textbf{Primitive I: Stable Pick-and-Place.}
Triggered by the centroid anchor, this primitive executes a canonical trapezoidal velocity profile. The manipulator approaches the anchor point $\mathbf{p}_{world}$ in the world coordinate system along the surface normal. Since the grasp point is localized near the geometric centroid, this strategy inherently maximizes stability through a balanced torque distribution.

\textbf{Primitive II: Pivot-based Reorientation.}
This primitive utilizes the workbench surface as an external pivot, with its dynamical evolution strictly governed by quasi-static moment equilibrium under frictional contact constraints. The execution is divided into four sequential phases:

\begin{enumerate}
    \item \textbf{Tangential Approach.} The system computes the edge tangent vector $\vec{t}$ based on the rim anchor. The arm then approaches the object's edge at a pre-tilt angle $\theta_{pre}$ along the tangential direction.
    \item \textbf{Pivoting Grasp.} The gripper closes to establish rigid contact, forming the initial contact configuration.
    \item \textbf{Leverage Rotation.} This stage represents the core dynamical process. During flipping, the contact point $C$ is treated as the instantaneous center of rotation (ICR), necessitating a no-slip velocity constraint:
    \begin{equation}
    v_{contact} = \mathbf{J}_C \dot{q} \approx 0
    \end{equation}
    where $\mathbf{J}_C$ denotes the contact Jacobian. At this juncture, the system dynamics satisfy the following torque inequality:
    % \begin{equation}
    % \boldsymbol{\tau}_{net} = \boldsymbol{\tau}_{arm} + \underbrace{(\mathbf{r}_{pivot} \times \mathbf{F}_{reaction})}_{\text{Support Torque}} - \underbrace{(\mathbf{r}_{com} \times m\mathbf{g})}_{\text{Gravity Torque}} > 0
    % \end{equation}
    \begin{equation}
    \begin{split}
    \boldsymbol{\tau}_{net} = \boldsymbol{\tau}_{arm} &+ \underbrace{(\mathbf{r}_{pivot} \times \mathbf{F}_{reaction})}_{\text{Support Torque}} \\
    &- \underbrace{(\mathbf{r}_{com} \times m\mathbf{g})}_{\text{Gravity Torque}} > 0
    \end{split}
    \end{equation}
    As formulated in the expression above, the reaction torque provided by the tabletop effectively counterbalances a significant portion of the gravitational torque. Consequently, the manipulator requires only a minimal active torque $\boldsymbol{\tau}_{arm}$ to break the equilibrium and achieve a stable, controlled flip.
    \item \textbf{Retract.} The arm lifts the object vertically to disengage contact constraints with the supporting surface, thereby completing the pose transition.
\end{enumerate}

\subsection{Few-Shot Adaptation Pipeline}
To address the demand for rapid industrial reconfiguration, Agile-VLA incorporates an efficient few-shot adaptation pipeline that enables minute-level deployment for previously unseen objects without requiring extensive retraining.

\textbf{Protocol-Driven Data Annotation.}
When introducing a novel workpiece, the system requires only $N=5$ labeled images. This protocol constructs a sparse supervisory dataset $\mathcal{D}_{few} = \{ (I_i, \mathbf{p}_i) \}_{i=1}^N$, where geometric priors implicitly define the mapping function $\Phi$. 

\textbf{Edge-Side Online Fine-Tuning.}
The perception head is updated locally on the Jetson Orin Nano. The optimization objective focuses on minimizing the anchor regression loss:
\begin{equation}
\mathcal{L}_{adapt} = \sum_{(I, \mathbf{p}) \in \mathcal{D}_{few}} \| \mathcal{F}_{vis}(I; \theta) - \mathbf{p} \|^2
\end{equation}
By reducing the complex manipulation task to geometric coordinate regression on a low-dimensional manifold, the training converges within minutes, effectively meeting industrial minute-level changeover requirements.

\textbf{Asynchronous Closed-Loop Execution.}
Following deployment, the perception stream (10Hz) and the control stream (50Hz) operate concurrently. A soft-synchronization mechanism facilitates dynamic switching between stable pick-and-place operations and pivot-based reorientation, ensuring robust, high-frequency closed-loop control regardless of edge-compute limitations.

\section{Experiments and Analysis}
This section presents a comprehensive real-world evaluation of Agile-VLA through physical deployment. The experiments are designed to quantitatively address four core research questions (RQs): \textbf{computational agility} (RQ1), large-scale task \textbf{generalization} and benchmark evaluation against open-source baselines (RQ2), dynamic \textbf{stability} and physical \textbf{robustness} (RQ3), and adaptation and architectural \textbf{ablation studies} (RQ4).

\subsection{Experimental Setup}
\textbf{Hardware Configuration.} The complete system is deployed on the resource-constrained edge platform NVIDIA Jetson Orin Nano (8GB). Execution is performed using a 6-DoF manipulator equipped with a Robotiq 2F-85 parallel gripper and a 6-axis force/torque sensor for recording dynamic interaction signals. Perception is provided by a fixed-view RGB-D camera, Intel RealSense D435i, which supplies global visual observations.

\textbf{Diverse Industrial Dataset (DID-127).} To overcome the limited test diversity in prior studies, a real-world benchmark termed DID-127 is constructed, comprising 127 industrial parts with diverse geometries and physical properties. Based on geometric topology and physical characteristics, the data set is divided into two subsets: easy subsets (75 types) and hard subsets (52 types). In total, more than 600 physical rollouts are executed.

\textbf{Open-Source Baselines.} To ensure fairness and scientific rigor, five representative and fully open-source embodied models are selected as baselines:Transporter (CoRL 2020), RT-1 (2022), RVT (CoRL 2023), Octo-Base (RSS 2024), OpenVLA (ICRA 2025). These models span diffusion-based policies, transformer-based visuomotor architectures, and large vision-language-action (VLA) models.

\subsection{Edge-Side Computational Agility Analysis (RQ1)}
A controlled system-level profiling study is first conducted on identical edge hardware to evaluate resource consumption and control fidelity.

\begin{table*}[t]
\centering
\renewcommand{\arraystretch}{1.1}
\setlength{\tabcolsep}{15pt}
\begin{tabular}{llcccc}
\toprule
Category & Method & $T_{\mathrm{inf}}$ (ms) $\downarrow$ & $f_{\mathrm{ctrl}}$ (Hz) $\uparrow$ & VRAM (GB) $\downarrow$ & TCP Jitter (mm) $\downarrow$ \\
\midrule
Sequence & RT-1 & $125 \pm 12$ & 8 & 3.8 & 3.25 \\
Diffusion & Octo-Base & $245 \pm 18$ & 4 & 3.2 & 4.12 \\
LLM-VLA & OpenVLA (4-bit) & $835 \pm 42$ & 1.2 & 6.8 & 5.82 \\
Ours & \textbf{Agile-VLA (Sync)} & $\mathbf{92 \pm 6}$ & 10.8 & \textbf{1.8} & 1.45 \\
Ours & \textbf{Agile-VLA (Async)} & $\mathbf{92}$ (Vis) & \textbf{50} & 2.1 & \textbf{0.65} \\
\bottomrule
\end{tabular}
\caption{System profiling on Jetson Orin Nano (8GB).}
\label{table:profiling}
\end{table*}

Table \ref{table:profiling} reveals the practical compute wall encountered by large-scale models under edge deployment. Even with 4-bit quantization, OpenVLA exhibits an autoregressive inference latency of approximately 835 ms, reducing the effective control frequency to 1.2 Hz and inducing high-frequency oscillations of 5.82 mm at the tool center point (TCP) during physical contact.

By contrast, Agile-VLA decouples perception and control through an asynchronous dual-stream design. The vision stream occupies only 2.1 GB of VRAM, while the control stream leverages cubic spline interpolation to elevate the execution frequency to servo-level 50 Hz. As a result, TCP jitter is reduced to 0.65 mm, fundamentally ensuring interaction agility at the hardware-constrained edge.

\subsection{Comprehensive Comparison with Open-Source SOTA Methods (RQ2)}
To quantitatively evaluate the impact of high-latency large models on closed-loop control, end-to-end pose reorientation experiments are conducted on the DID-127 dataset. In addition to task success rate, two additional metrics are introduced to evaluate execution quality: collision rate and the low-level kinematic indicator jerk ($m/s^3$).

\begin{table*}[t]
\centering
\renewcommand{\arraystretch}{1.1}
\setlength{\tabcolsep}{10pt}

\begin{tabular}{lcccccc}
\toprule
Method 
& Easy SR (\%) $\uparrow$ 
& Hard SR (\%) $\uparrow$ 
& Avg SR (\%) $\uparrow$ 
& Time (s) $\downarrow$ 
& Collision (\%) $\downarrow$ 
& Jerk (m/s$^3$) $\downarrow$ \\
\midrule

Transporter & 74.5 & 28.6 & 55.4 & 11.2 & 16.5 & 10.21 \\
RT-1 & 78.2 & 34.0 & 60.1 & 14.5 & 18.2 & 12.84 \\
RVT & 82.5 & 36.5 & 63.8 & 12.8 & 14.2 & 11.56 \\
Octo-Base & 85.0 & 38.5 & 61.7 & 16.8 & 15.4 & 14.35 \\
OpenVLA (4-bit) & 88.6 & 46.5 & 71.0 & 19.5 & 21.8 & 15.62 \\
\rowcolor{gray!15}
\textbf{Agile-VLA} 
& \textbf{96.5} 
& \textbf{84.6} 
& \textbf{90.5} 
& \textbf{8.4} 
& \textbf{3.2} 
& \textbf{1.24} \\
\bottomrule
\end{tabular}
\caption{Multi-dimensional performance on DID-127 dataset.}
\label{table:4.3}
\end{table*}

As shown in Table \ref{table:4.3}, end-to-end large-scale models (e.g., OpenVLA, Octo) demonstrate reasonable semantic understanding on regular industrial parts. However, under edge deployment constraints, extreme inference latency induces a pronounced open-loop blind actuation phenomenon during physical execution. Empirical measurements indicate that such latency leads to excessive motion jerk (Jerk $>$ 14 $m/s^3$) and hazardous collision rates ($>$ 15\%), resulting in a sharp degradation in success rate when handling objects with complex topology.

In contrast, Agile-VLA leverages asynchronous decoupling to factor high-dimensional visual reasoning into a lightweight geometric manifold projection. This mechanism reduces the overall execution time to 8.4 s while utilizing a 50 Hz cubic spline interpolation kernel at the control layer to suppress motion jerk to 1.24 $m/s^3$. The results demonstrate that computational agility and smooth compliant interaction can be achieved simultaneously during pose rectification.

\subsection{Dynamic Validation of Dexterous Reorientation for Irregular Objects (RQ3)}
\begin{figure} 
\centering 
\includegraphics[width=1.0\linewidth]{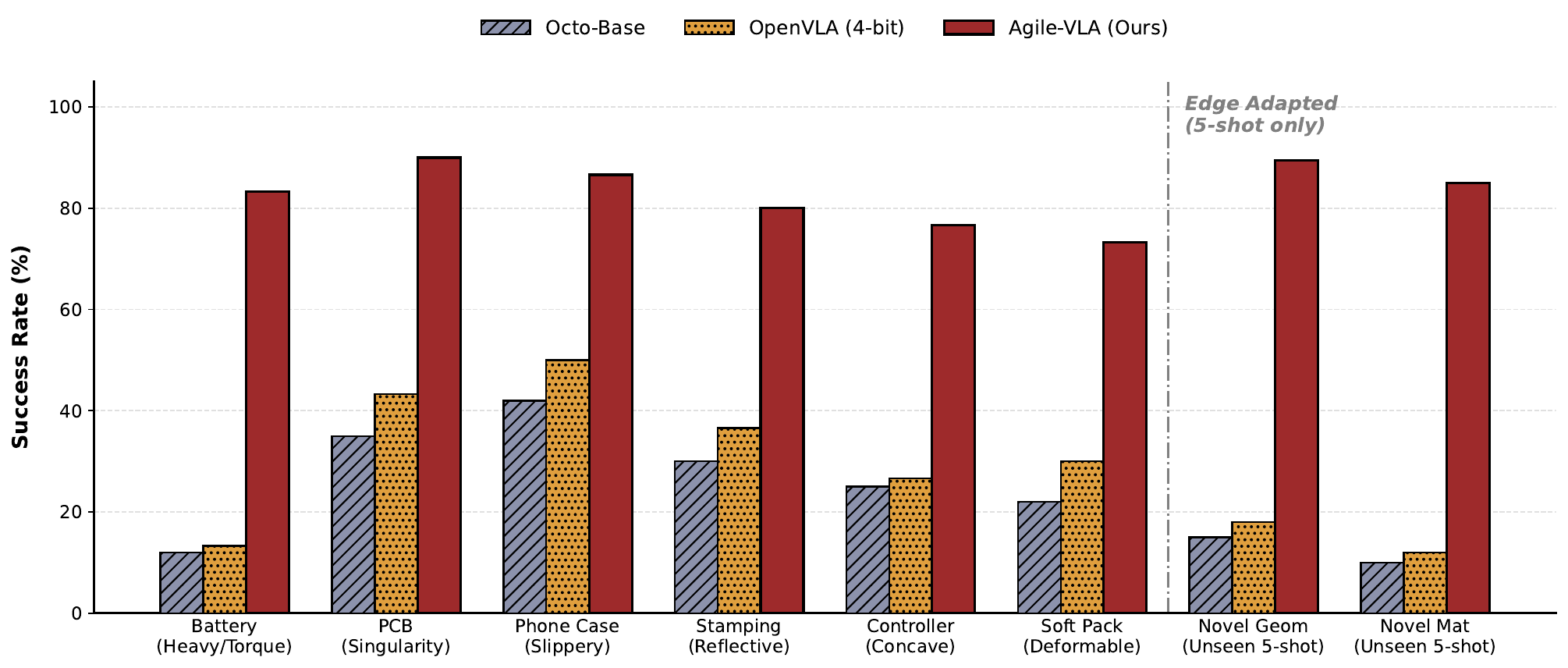} 
\caption{ \textbf{Real-world multi-task reorientation performance across extreme physical conditions.} We systematically benchmark Agile-VLA against state-of-the-art foundation models on six extreme industrial parts (involving heavy masses and complex topologies) and two novel out-of-distribution objects. \textit{Crucially}, the performance on 'Unseen' objects reflects the results after our 5-shot edge adaptation pipeline, demonstrating the framework's rapid deployment capability in contrast to the massive fine-tuning required by large VLA baselines. By leveraging extrinsic pivoting, Agile-VLA successfully bypasses actuator torque overload (e.g., Heavy Battery) and kinematic singularities (e.g., Complex PCB), achieving consistently higher success rates.} 
\label{fig:chart} 
\end{figure}

To address the kinematic limitations of 6-DoF manipulators in complex spatial reorientation, a comparative study is conducted to validate the proposed extrinsic dexterity mechanism. Figure \ref{fig:chart} illustrates the success rates of Agile-VLA compared to state-of-the-art foundation models across six extreme physical conditions and two unseen geometries.
The performance gap in Figure \ref{fig:chart} highlights the physical necessity of incorporating environmental constraints. Beyond success rates, two critical dynamic indicators were extracted at the flipping instant to explain the baseline failures: peak wrist torque ($\tau_{max}$) and singularity trigger rate.

% \begin{table}[htbp]
% \centering
% \renewcommand{\arraystretch}{1.1}
% \setlength{\tabcolsep}{6pt}
% \small

% \begin{tabular}{@{}ccccc@{}}
% \toprule
% Object & Strategy & SR (\%) $\uparrow$ & $t_{\max}$ (N$\cdot$m) $\downarrow$ & Sing. (\%) $\downarrow$ \\
% \midrule

% \multirow{2}{*}{Battery\textsuperscript{a}}
% & Baseline & 13.3 & 12.8 & 46.6 \\
% & \textbf{Ours} & \textbf{83.3} & \textbf{3.5} & \textbf{0} \\
% \cmidrule(lr){1-5}

% \multirow{2}{*}{PCB\textsuperscript{b}}
% & Baseline & 43.3 & 3.1 & 20 \\
% & \textbf{Ours} & \textbf{90} & \textbf{0.8} & \textbf{0} \\
% \cmidrule(lr){1-5}

% \multirow{2}{*}{Phone Case\textsuperscript{c}}
% & Baseline & 50 & 1.8 & 10 \\
% & \textbf{Ours} & \textbf{86.6} & \textbf{0.6} & \textbf{0} \\
% \cmidrule(lr){1-5}

% \multirow{2}{*}{Stamping\textsuperscript{d}}
% & Baseline & 36.6 & 5.8 & 26.6 \\
% & \textbf{Ours} & \textbf{80} & \textbf{1.5} & \textbf{0} \\
% \bottomrule
% \end{tabular}

% \caption{Kinematic \& dynamic evaluation on extreme objects.
% \newline
% \textsuperscript{a} 1.2 kg off-center mass.
% \textsuperscript{b} Topology highly prone to kinematic interference.
% \textsuperscript{c} Ultra-low surface friction.
% \textsuperscript{d} Highly reflective edges.
% }
% \label{table:extreme_eval}

% \end{table}

\textbf{Singularity Avoidance.} When a 6-DoF manipulator attempts to forcibly execute a 180° aerial orientation interpolation for complex topologies (e.g., the PCB), the resulting trajectory frequently traverses kinematic singularities, with trigger rates reaching 20\%–46\% for baseline models. In contrast, Agile-VLA's boundary-anchored pivot strategy reduces high-dimensional spatial rotation to a two-dimensional linkage motion, fundamentally eliminating singularity traversal and reducing the trigger rate to 0.0\%.

\textbf{Dynamic Offloading.} During the aerial flipping of a heavy eccentric object (Battery, 1.2 kg), the rapid increase in lever arm length causes the peak wrist torque to surge to 12.8 N$\cdot$m, approaching the safety limit of collaborative manipulators. Agile-VLA rigorously satisfies the moment equilibrium condition, utilizing the workbench's support torque to reduce the required actuation torque by 72.6\% to just 3.5 N$\cdot$m. These physical metrics provide strong empirical evidence that external environmental constraints are indispensable for stable and efficient edge-deployed manipulation systems.

\subsection{Few-Shot Adaptation and Architectural Ablation (RQ4)}

\begin{table}[t]
\centering
\renewcommand{\arraystretch}{1.1}
\setlength{\tabcolsep}{6pt}
\small

\begin{tabular}{@{}c p{1.8cm} c p{1.8cm}@{}}
\toprule
Samples ($N$) 
& Pixel Error (px) $\downarrow$ 
& Avg SR (\%) $\uparrow$ 
& Tuning Time (min) $\downarrow$ \\
\midrule

$N = 1$  & 18.4 & 35 & 0.8 \\

\rowcolor{gray!12}
\textbf{$N = 5$}  
& \textbf{3.5} 
& \textbf{90} 
& \textbf{2.5} \\

$N = 10$ & 3.2 & 90 & 5.1 \\
\bottomrule
\end{tabular}
\caption{Impact of annotation samples ($N$).}
\label{4.5:IV}
\end{table}

To evaluate deployment agility in real-world industrial environments, Table \ref{4.5:IV} illustrates the convergence efficiency of the perception projection layer when fine-tuning on previously unseen novel, irregular objects. Owing to implicit affordance anchoring, complex action decoding is reduced to low-dimensional point regression on a geometric manifold, substantially decreasing data dependency. With only five sparsely annotated samples, the model converges within 2.5 minutes to a prediction error of 3.5 px and achieves a task success rate of 90.0\%, satisfying the practical requirement of minute-level production line switching.

\begin{table}[t]
\centering
\renewcommand{\arraystretch}{1.1}
\setlength{\tabcolsep}{5pt}
\small

\begin{tabular}{ccccc}
\toprule
Async 
& Anchor 
& Pivot 
& SR (\%) $\uparrow$ 
& Jitter (mm) $\downarrow$ \\
\midrule

$\times$ (Sync) 
& $\checkmark$ 
& $\checkmark$ 
& 60.5 
& 1.45 \\

$\checkmark$ 
& $\times$ (Center) 
& $\checkmark$ 
& 43.2 
& 0.68 \\

$\checkmark$ 
& $\checkmark$ 
& $\times$ (Direct) 
& 18.5 
& 0.82 \\

\rowcolor{gray!12}
$\checkmark$ 
& $\checkmark$ 
& $\checkmark$ 
& \textbf{90.5} 
& \textbf{0.65} \\

\bottomrule
\end{tabular}

\caption{Ablation study on core architecture components.}
\label{4.5:V}

\end{table}

Table \ref{4.5:V} further quantifies the independent contributions of core architectural components through structured ablation experiments. Removing the external pivot constraint results in the system's inability to overcome kinematic limitations, causing the success rate to plummet to 18.5\%. Replacing implicit anchoring with a conventional global-centroid strategy introduces severe physical interference during flipping, reducing the success rate to 43.2\%. In addition, eliminating the asynchronous dual-stream architecture exposes perception-layer latency, doubling TCP jitter (to 1.45 mm) and increasing the likelihood of contact collisions.

\section{Conclusion}
This paper presents Agile-VLA, a lightweight hybrid control framework specifically tailored for complex industrial pose reorientation under compute-constrained edge deployment. To address the fundamental limitations of existing end-to-end VLA models regarding the trade-off between high-frequency real-time control and manipulation dexterity on edge hardware, an asynchronous dual-stream decoupling architecture and an implicit affordance anchoring mechanism are introduced.

By implicitly projecting high-dimensional language semantics into geometric keypoint representations and leveraging environmental contact constraints to activate extrinsic dexterity in a standard 6-DoF manipulator, stable closed-loop execution is achieved on a single NVIDIA Jetson Orin Nano (8GB), sustaining 10 Hz semantic perception alongside 50 Hz dynamic control. Across an industrial-scale evaluation involving 127 real-world components, Agile-VLA demonstrates strong physical consistency. A state-of-the-art success rate of 90.5\% is achieved on objects with complex topology, while required actuation torque and collision incidence are reduced by 72.6\% through principled dynamic optimization. With only five annotated samples, adaptation to new objects is completed within three minutes, highlighting minimal data dependency and practical deployment efficiency.

Future work will investigate deeper multimodal integration—particularly the incorporation of high-frequency tactile feedback constraints—to further advance robustness under dynamic disturbances and extreme manipulation scenarios.

\addtolength{\textheight}{-12cm}   % This command serves to balance the column lengths

%%%%%%%%%%%%%%%%%%%%%%%%%%%%%%%%%%%%%%%%%%%%%%%%%%%%%%%%%%%%%%%%%%%%%%%%%%%%%%%%

%%%%%%%%%%%%%%%%%%%%%%%%%%%%%%%%%%%%%%%%%%%%%%%%%%%%%%%%%%%%%%%%%%%%%%%%%%%%%%%%

%%%%%%%%%%%%%%%%%%%%%%%%%%%%%%%%%%%%%%%%%%%%%%%%%%%%%%%%%%%%%%%%%%%%%%%%%%%%%%%%

%%%%%%%%%%%%%%%%%%%%%%%%%%%%%%%%%%%%%%%%%%%%%%%%%%%%%%%%%%%%%%%%%%%%%%%%%%%%%%%%

\end{document}